\documentclass[runningheads]{llncs}
\usepackage{graphicx}
\usepackage{subcaption}
\usepackage{xcolor}
\usepackage{amsmath}
\usepackage[T1]{fontenc}
\usepackage{graphicx}
\usepackage{hyperref}

\begin{document}
\title{Federated Discrete Denoising Diffusion Model for Molecular Generation with OpenFL}


\author{Kevin Ta \inst{1} \and
Patrick Foley \inst{1} \and
Mattson Thieme \inst{2} \and
Abhishek Pandey \inst{2} \and
Prashant Shah \inst{1}
}

\institute{
Intel Corporation, Santa Clara, CA 95052, USA \and
AbbVie Inc., North Chicago, IL 60064, USA}
\maketitle              
\begin{abstract}
Generating unique molecules with biochemically desired properties to serve as viable drug candidates is a difficult task that requires specialized domain expertise. In recent years, diffusion models have shown promising results in accelerating the drug design process through AI-driven molecular generation. However, training these models requires massive amounts of data, which are often isolated in proprietary silos. OpenFL is a federated learning framework that enables privacy-preserving collaborative training across these decentralized data sites. In this work, we present a federated discrete denoising diffusion model that was trained using OpenFL. The federated model achieves comparable performance with a model trained on centralized data when evaluating the uniqueness and validity of the generated molecules. This demonstrates the utility of federated learning in the drug design process. \\

OpenFL is available at: \href{https://github.com/securefederatedai/openfl}{https://github.com/securefederatedai/openfl}

\end{abstract}
\section{Introduction}
The discovery and development of novel therapeutics is a resource-intensive process that requires deep domain expertise. Generative machine learning models have emerged as powerful tools for drug discovery and have demonstrated the potential to generate molecules with pharmacologically viable properties \cite{Loeffler2024-zw,decao2022molganimplicitgenerativemodel,Zhavoronkov2019-kq,vignac2023digress}. This can ultimately lead to more efficient drug development. However, the performance and generalizability of machine learning models are heavily dependent on the amount of data available, which is often siloed across different research institutions and pharmaceutical companies. Combining all of this data can capture a more comprehensive and representative distribution of features, leading to a more robust model. However, this is not feasible due to privacy and legal concerns, competitive pressure, and technical constraints.

Federated Learning (FL) has emerged as a promising solution to this challenge in the healthcare industry \cite{PATI2024100974,Rieke2020-iv,Sheller2020-pl}. FL is a collaborative machine learning paradigm that enables multiple clients, such as research institutions or companies, to jointly train a shared global model without directly exchanging their locally stored data. Additionally, confidential computing, which involves the use of hardware-based technologies to create secure enclaves for data processing, ensures that data remains encrypted during processing. This provides an additional layer of security and trust. A decentralized machine learning approach combined with confidential computing can help mitigate many systemic privacy risks associated with traditional centralized learning (CL) while enabling collaborative model development that will benefit all clients involved. Open Federated Learning (OpenFL) \cite{Foley2022-vl} is an open source framework developed by Intel specifically designed to train and evaluate machine learning algorithms in a more secure, collaborative manner across decentralized data sites.

In this work, we introduce a novel framework for training a class of generative models called a discrete denoising diffusion model using OpenFL. This approach enables collaborative training on decentralized datasets of molecules while simultaneously preserving data privacy and security, allowing for a more robust and comprehensive molecular generation model for the discovery of novel drug candidates.

\section{Federated Learning and OpenFL}

Federated Learning allows multiple parties to collaboratively train a machine learning model without sharing their private data \cite{McMahanMRA16}. Instead of sending raw data to a central server, each client trains a local model on their respective data and sends the model updates to a central server for aggregation. This aggregated global model is then shared back with the clients where they will resume training on their local models, iteratively improving the model's performance using the collective data.

The OpenFL framework uses a client-server architecture, where a central server is responsible for aggregating and distributing the latest global model weights and coordinating tasks across all connected clients. A model owner is responsible for designing the general experiment, including the model architecture, training and validation sequence, and an overall plan that is agreed upon by all parties of the federation. An initial global model is sent to the client sites to train on the local data using the specifications and plan agreed upon for the federation. Once training is complete, the update model is sent to the aggregator server, which then combines these into a new global model.

OpenFL is built with a security-first approach, which helps ensure that the intellectual property of the global model is protected throughout the training process. Communication between the aggregator node and the collaborator nodes occurs through a mutually authenticated transport layer security (mTLS) connection. This channel safeguards the transmitted model updates from unauthorized access. Additionally, OpenFL is designed to run efficiently in trusted execution environments (TEE), which are secure areas within the processor that can isolate code and data from other operations. This adds a layer of confidentiality and prevents users from copying the model or data out of the TEE. A TEE also enables remote attestation for users to verify the integrity of the code being executed. TEEs are a fundamental component of confidential computing, ensuring that sensitive data and computations are protected from unauthorized access and tampering. OpenFL further restricts information that is shared between collaborators and aggregators to the model weights and aggregated metrics, ensuring that the training data and information about the data remain localized to the collaborators. While no product or component can be absolutely secure, by combining federated learning with confidential computing, this approach adheres to the principle of data minimization, greatly reducing the risk of data leaks and privacy violations.

\section{Molecular Generation and Denoising Diffusion Models}
Molecular generation models aim to learn the underlying distribution of molecules in a given latent space and generate novel molecules with desired biochemical properties. These models learn from existing molecular data to capture complex relationships between molecular structure and its properties. The generated molecules can then be screened by domain experts to evaluate the validity of molecules to potentially be further used for drug discovery.

Denoising diffusion models are a powerful class of generative models that work by gradually deconstructing a data distribution through the iterative introduction of noise \cite{ho2020denoising}. The model is then trained to reverse this process, learning to progressively denoise the data and construct a new sample from the learned distribution. In the context of molecular generation, a random noise distribution is iteratively refined until valid and unique molecules are generated. DiGress \cite{vignac2023digress} is a specific type of denoising diffusion model designed to generate molecules in the form of graphs. DiGress represents molecules as graphs, where the graph nodes represent atoms and the graph edges represent the bonds, and learns to generate new molecules by iteratively denoising the graph starting from a noisy distribution. The DiGress model also introduces a novel discrete guidance mechanism. Through this mechanism, an additional regressor is independently trained to predict the target properties of a graph from the noisy graph and guide the generation toward graphs with these desired properties during inference. However, the original DiGress model was designed for centralized training, which may be incomplete in the context of drug discovery where a large portion of data is isolated across different companies and institutions. 

\section{Federated Denoising Diffusion Model with OpenFL}
We present a framework for training DiGress, a discrete denoising diffusion model, in a federated setting using OpenFL to leverage the sensitive data siloed across the pharmaceutical industry. This approach aims to combine the benefits of privacy-preserving FL with the generative capabilities of DiGress for molecular generation. 

We preserve the training algorithm described in the original DiGress paper \cite{vignac2023digress}. Starting with an input graph, noise is iteratively added to the nodes and edges. Spectral and structural features are then computed at each iteration. The noisy graph, along with the spectral and structural features, are passed through the denoising model. The denoising model is trained to minimize the cross-entropy loss between the denoised nodes and edges and the original input graph. In parallel, a regressor for conditional generation is trained to predict the desired properties of a clean graph from a noisy graph by minimizing the mean squared error between the predicted properties and target properties. For this work, the regressor trains to predict the highest occupied molecular orbital and dipole moment $\mu$. To federate this training scheme, OpenFL offers a highly flexible and customizable Workflow API.

\begin{figure}[t]
\includegraphics[width=\textwidth]{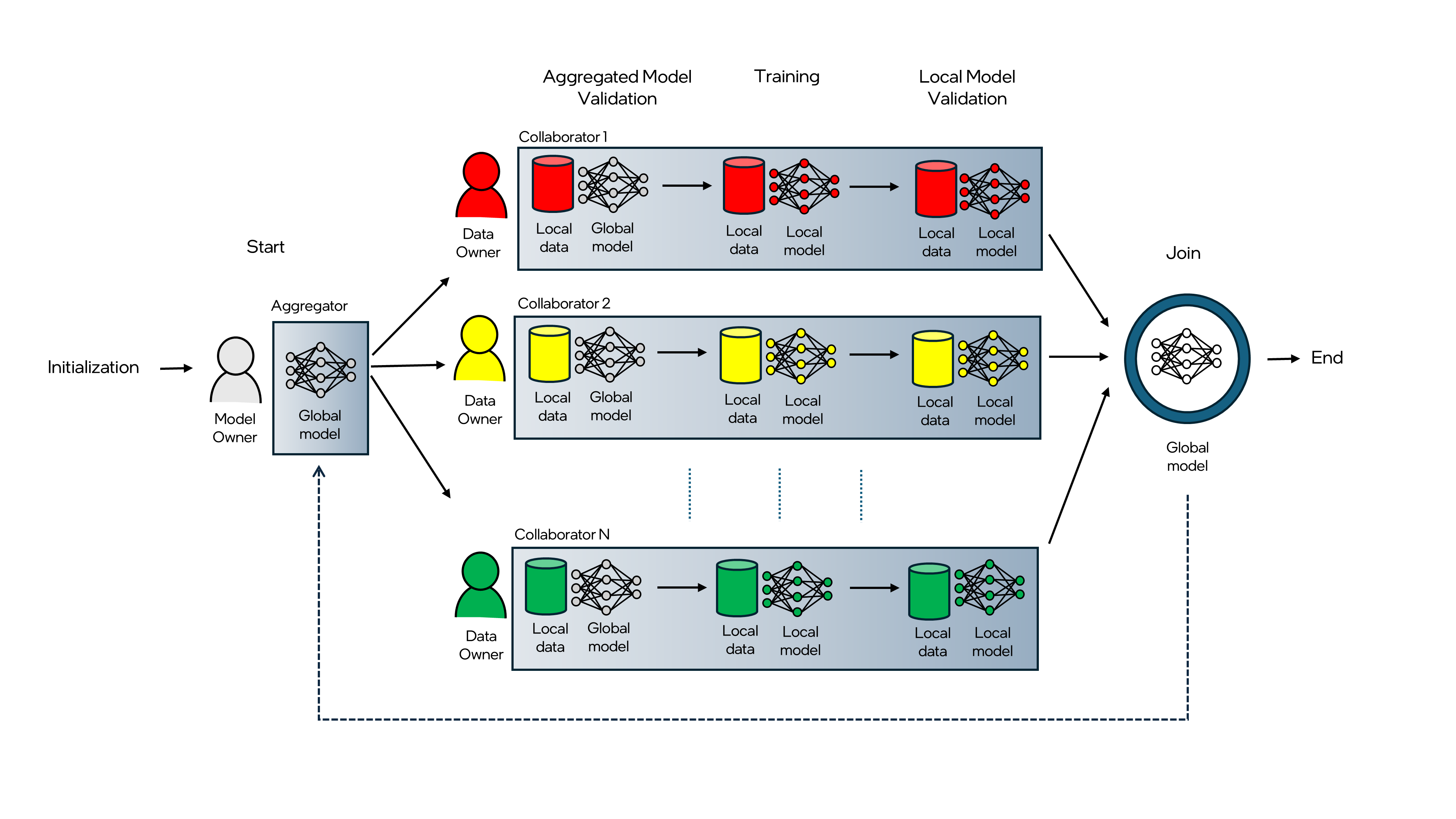}
\caption{\small OpenFL Workflow API: A global model is sent to collaborators for validation and local training. Each collaborator validates and updates the model with local data, then sends the updated model back to the aggregator. The aggregator combines these updates to form a new global model, completing one federation round.} \label{fig1}
\end{figure}

The Workflow API is an experimental interface that offers significant flexibility for researchers to set up and execute federated learning experiments within a simulation environment. Integrating DiGress's denoising and regressor models into this interface requires only minor modifications to the original codebase. The workflow is defined as a sequence of tasks that can be assigned to execute either at the aggregator or the collaborator site during runtime. Users can control the information passed through each task and define custom training, validation loops, and federated aggregation algorithms. The general sequence is as follows:
\begin{itemize}
  \item \textbf{Flow Definition}: Every workflow begins with a start task and concludes with an end task. The flow is defined as a subclass of OpenFL's \textit{FLSpec} class, which specifies the sequence of tasks and their execution logic.
  \item \textbf{Initialization}: The workflow is then initialized with several customizable parameters, including the initial state dictionaries and optimizers for both the regressor and denoising models, as well as the number of training rounds. These parameters are stored as instance variables to be used throughout the workflow.
  \item \textbf{Start}: The aggregator site starts the workflow and prepares to send the global model and tasks to the list of collaborators in the federation.
  \item \textbf{Aggregated Model Validation}: The data owner at the collaborator sites receive the global model and performs validation on their local data and report the validation metrics. Once validation is complete, the collaborators transition to the train task
  \item \textbf{Training}: The collaborators train the model on their local data. The training loss is reported and the model updates are stored. After training, the collaborators transition to the local model validation task.
  \item \textbf{Local Model Validation}: The collaborator sites rerun validation on the newly trained local model to assess performance with the local model updates and report the validation metrics. Once validation is complete, the collaborators send the model updates and metrics back to the aggregator.
  \item \textbf{Join}: The aggregator executes the join task and aggregates the model updates to form a new, updated global model using either a built-in or custom aggregation function. For this work, we use weighted federated averaging \cite{McMahanMRA16}, which is forumlated as follows:
\begin{equation}
w_{t+1} = \sum_{k=1}^K \alpha_k w_t^k
\end{equation}
where \( K \) is the number of clients, \( w_t^k \) are the model parameters from client \( k \) at iteration \( t \), and \( \alpha_k \) are the weights for each client such that \( \sum_{k=1}^K \alpha_k = 1 \). The metrics are also averaged during this task. If the training is complete, the aggregator will transition to the end task, otherwise, the aggregator will send the global model back to collaborators and restart the sequence at the perform aggregated model validation task.
  \item \textbf{End}: The aggregator ends the experiment, marking the conclusion of the workflow.
\end{itemize}

In addition, for production-ready settings that require a more robust security posture (such as mTLS, TEE, etc.), DiGress can be seamlessly adapted to OpenFL's Task Runner API. This API is designed to support real-world federations and consists of short-lived components that terminate at the end of the experiment. The Task Runner API follows the same general sequence as the Workflow API but operates in the backend, allowing users to focus on the core data science tasks such as model definition, training/validation code, and data loaders. All training and validation logic, as well as data handling, are defined within a shareable workspace, and the overall federated learning plan is predefined and agreed upon by all participants in the experiment. In the next section, we report numbers from a feasibility study conducted using the Task Runner API.

\section{Experiment \& Results}

\begin{table}[t]
\caption{Comparing Evaluation Metrics Between CL and FL Experiments}\label{tab1}
\begin{tabular}{|l|l|l|l|l|}
\hline
Method & NLL (Diffusion) & MAE (Regressor) & Validity & Uniqueness \\
\hline
Centralized Learning (CL) & 68.45 & 0.6932 &  0.9600 & 0.9958 \\
\hline
Federated Learning (FL) & 70.58 & 0.7026 &  0.9560 & 0.9989 \\
\hline
Absolute \%-Difference & \bf{3.06} & \bf{1.35} & \bf{0.42} & \bf{0.31} \\
\hline
\end{tabular}
\smallskip 
\parbox{0.8\textwidth}{\footnotesize\textit{Note: For NLL (Diffusion) and MAE (Regressor), lower values are better. For Validity and Uniqueness, higher values are better.}}
\end{table}

\begin{figure}[t]
\centering
\begin{subfigure}{0.48\textwidth}
  \includegraphics[width=\linewidth]{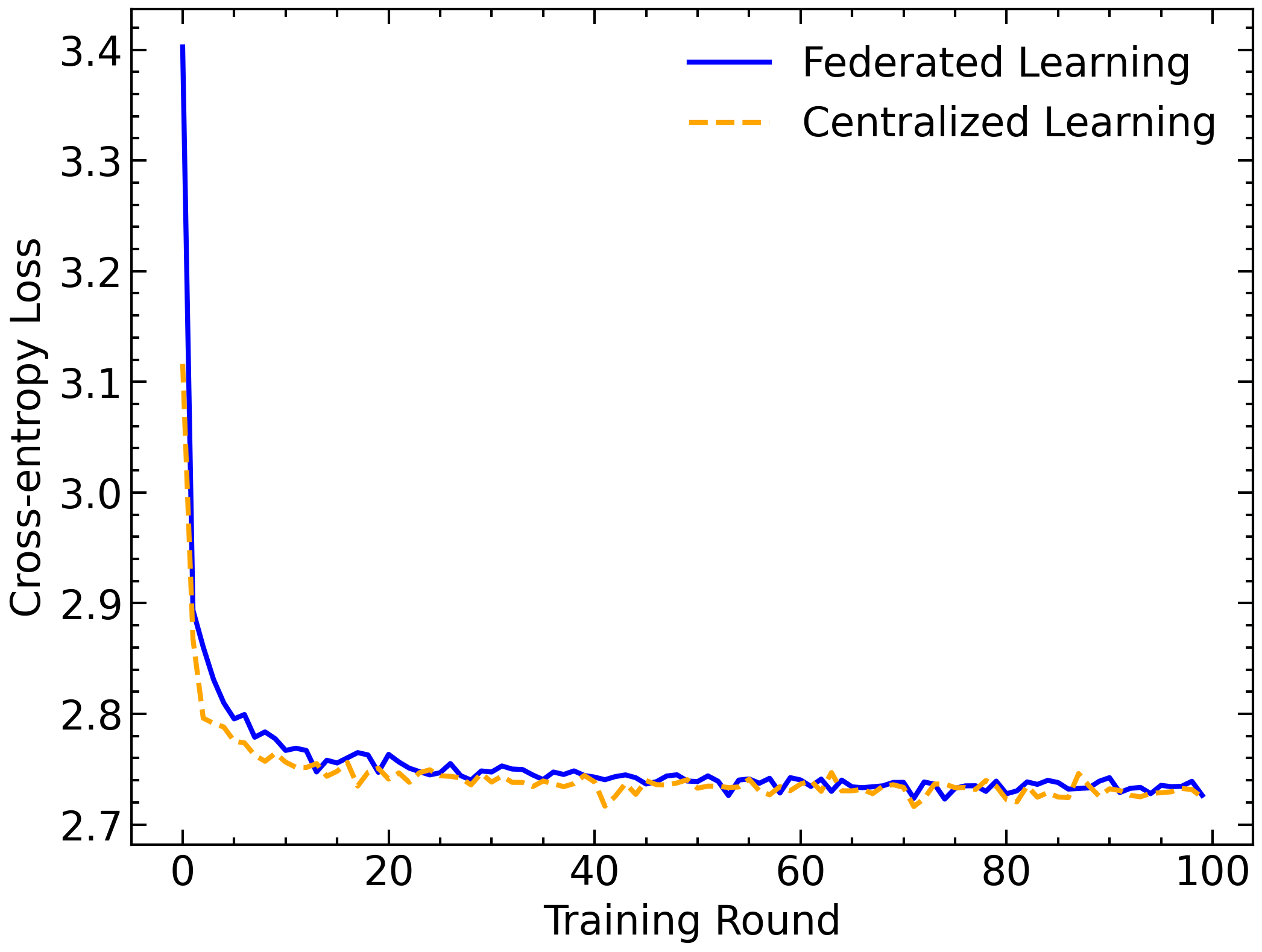}
  \caption{\small Denoising Model}
  \label{fig:train1}
\end{subfigure}
\hfill
\begin{subfigure}{0.48\textwidth}
  \includegraphics[width=\linewidth]{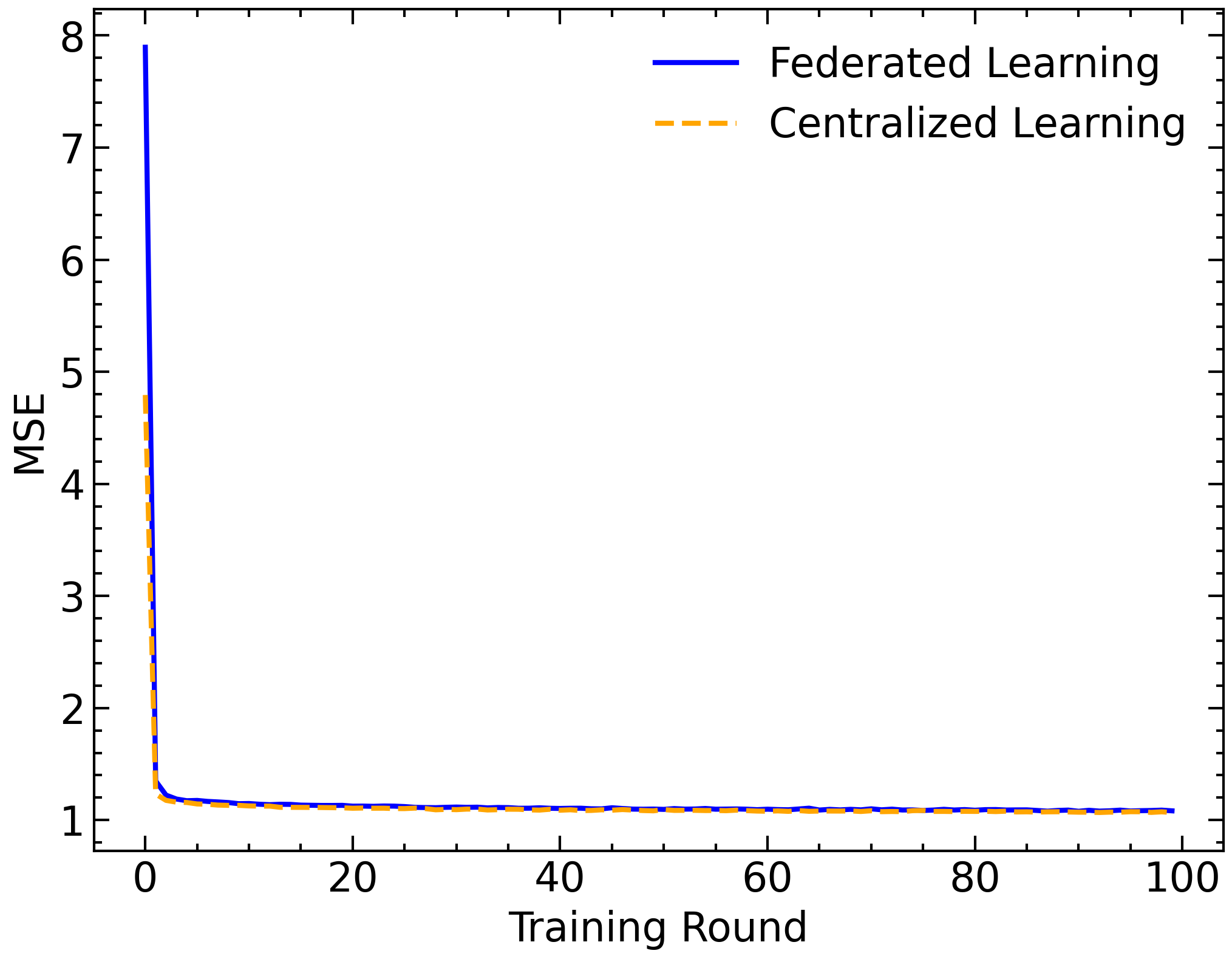}
  \caption{\small Regressor Model}
  \label{fig:train2}
\end{subfigure}
\caption{Model training loss across rounds}
\label{fig:training-curves}
\end{figure}

\begin{figure}[t]
\centering
\begin{subfigure}{0.48\textwidth}
  \includegraphics[width=\linewidth]{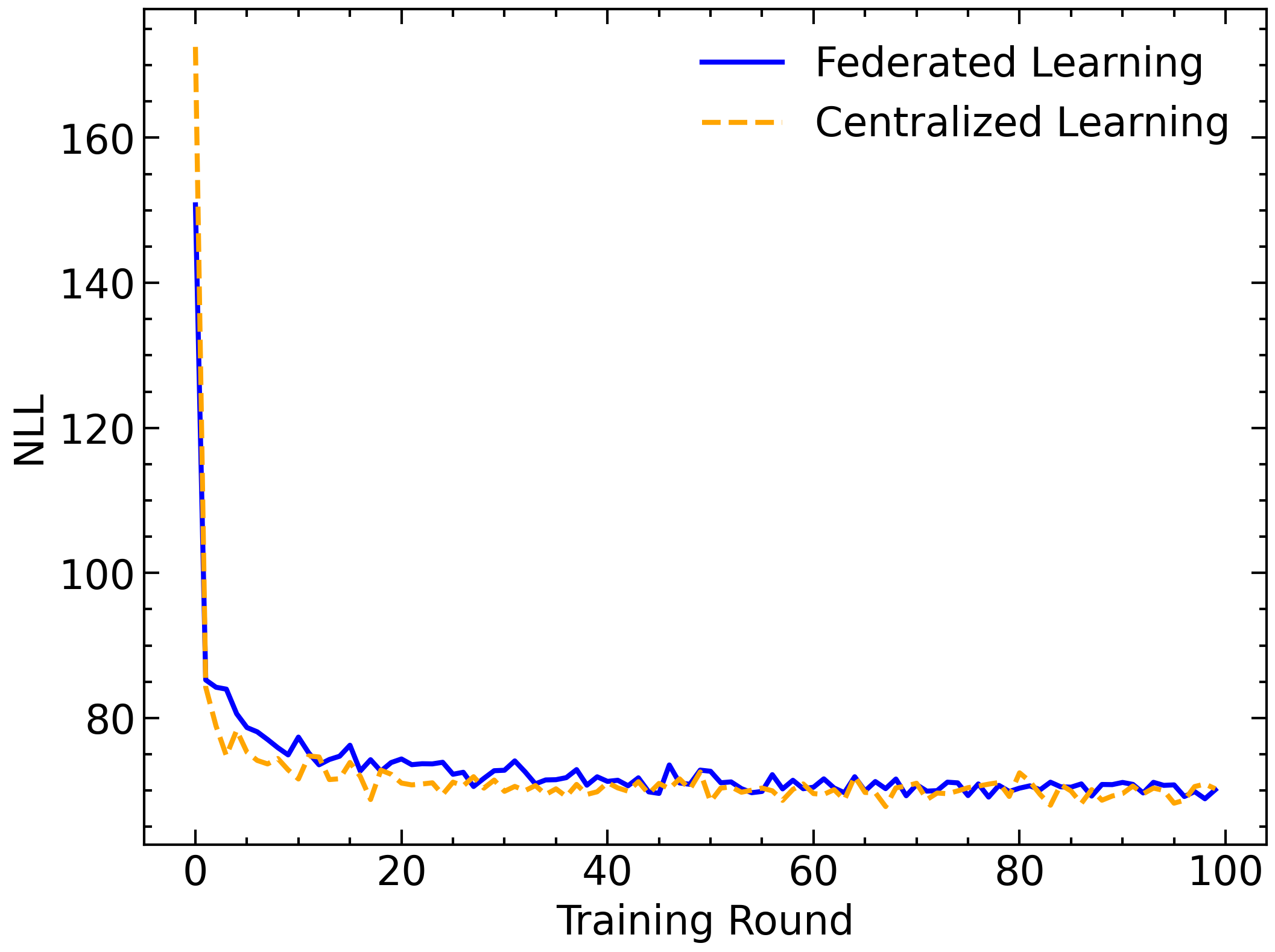}
  \caption{\small Denoising Model}
  \label{fig:val1}
\end{subfigure}
\hfill
\begin{subfigure}{0.48\textwidth}
  \includegraphics[width=\linewidth]{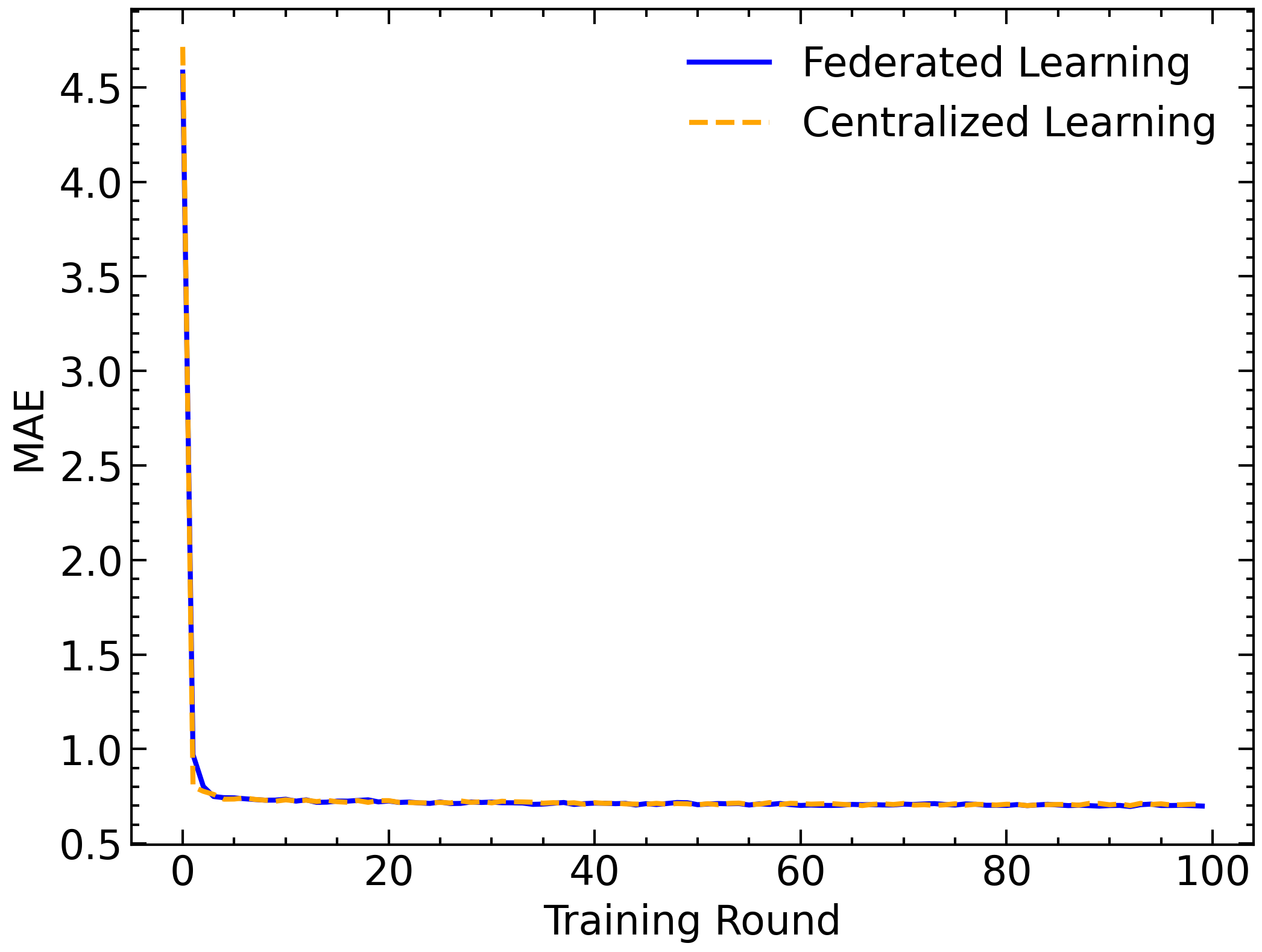}
  \caption{Regressor Model}
  \label{fig:val2}
\end{subfigure}
\caption{\small Model validation loss across rounds}
\label{fig:validation-curves}
\end{figure}

\begin{figure}[t]
\centering
\begin{subfigure}{0.48\textwidth}
  \includegraphics[width=\linewidth]{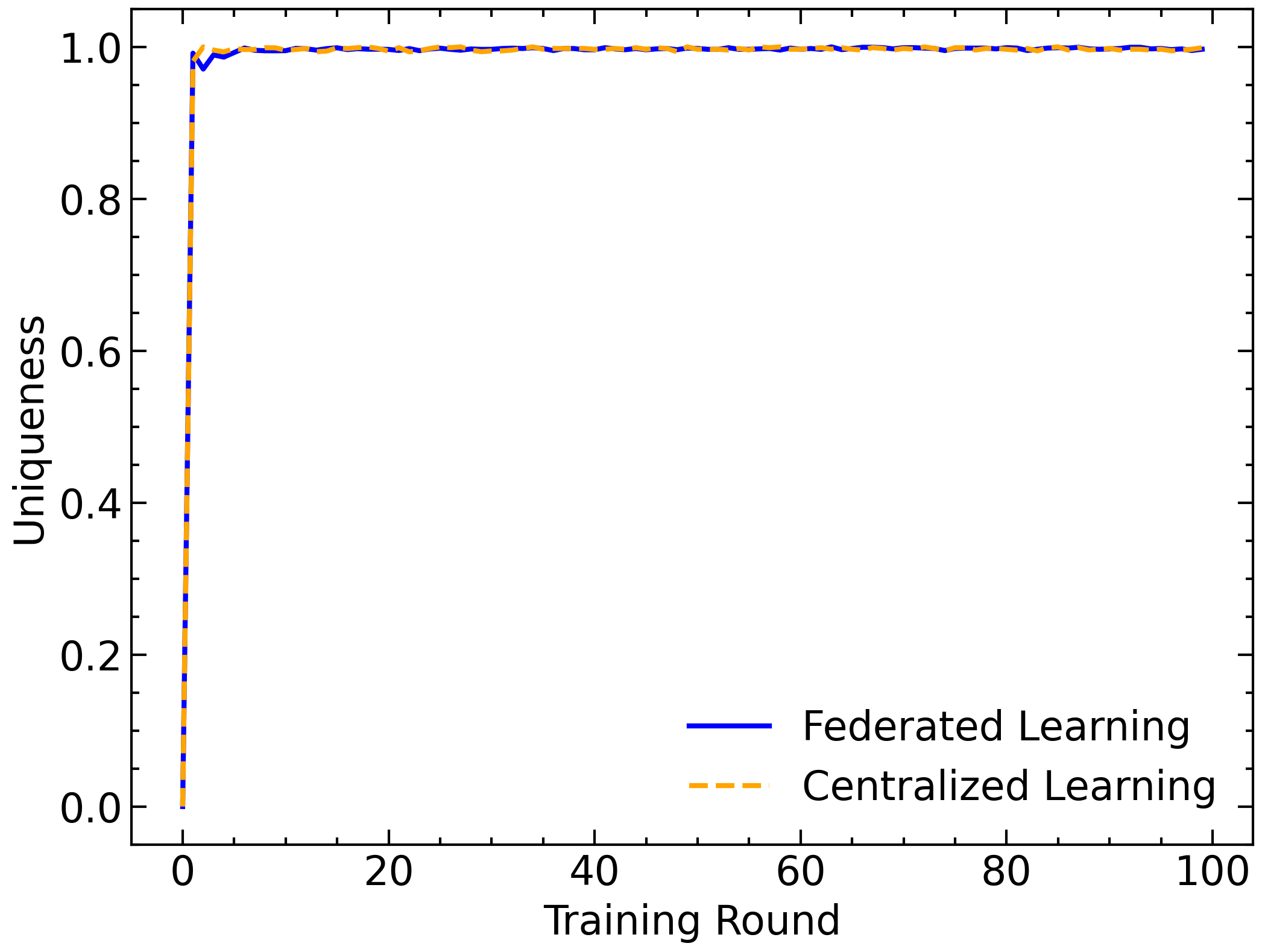}
  \caption{\small Uniqueness}
  \label{fig:samp1}
\end{subfigure}
\hfill
\begin{subfigure}{0.48\textwidth}
  \includegraphics[width=\linewidth]{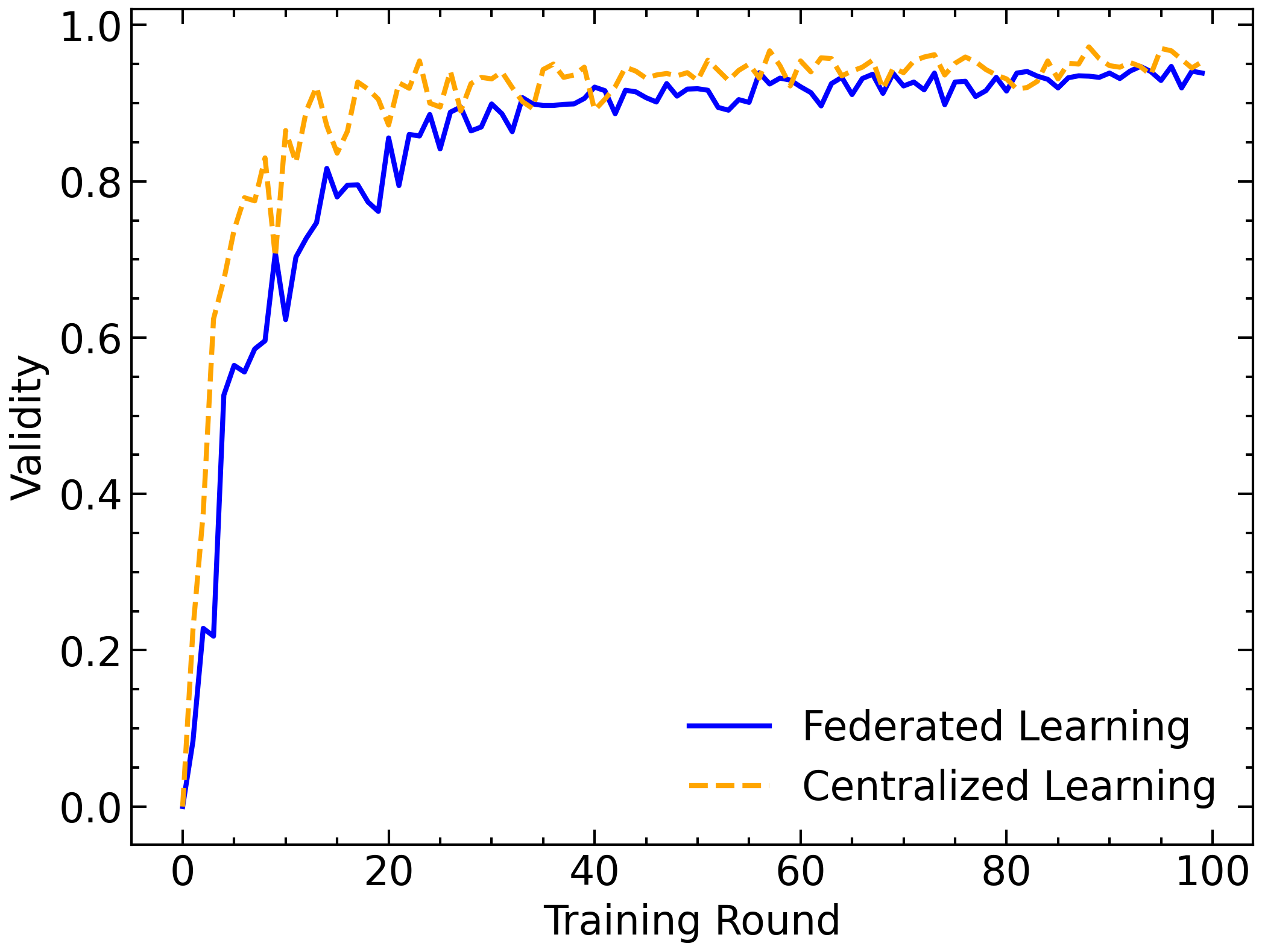}
  \caption{\small Validity}
  \label{fig:samp2}
\end{subfigure}
\caption{\small Sampling metrics across rounds}
\label{fig:sampling-metrics}
\end{figure}

To demonstrate the feasibility of FL for molecular generation, we utilize the QM9 dataset\cite{ramakrishnan2014quantum,Ruddigkeit2012}, which comprises of over 130k small organic molecules, each represented by its SMILES string and associated quantum chemical properties. This dataset is particularly suitable for our task due to its diversity and molecular validity. To simulate a federated learning environment, we randomly split the QM9 dataset across two collaborators. Each collaborator's data is further divided into 80\% for training, 10\% for validation, and 10\% for testing. For centralized learning, we employ the same 80/10/10 split on the entire dataset. This setup ensures that both learning paradigms are evaluated under comparable conditions

The model architecture consists of a denoising diffusion model and a regressor model that are trained concurrently, but separately. Both models are trained with a batch size of 512 using an AdamW optimizer with a learning rate of 2e-4 and a weight decay of 1.0e-12. During training and evaluation, the model generates 1,000 samples using 100 diffusion steps across 10 diffusion chains. For FL, we train the model for 100 rounds, with each round consisting of 1 epoch per collaborator. In contrast, the CL model is trained for 100 epochs. This approach ensures that both models undergo an equivalent number of training iterations, allowing for a fair comparison.

We evaluate the models using four key metrics: Negative Log-Likelihood (NLL) for the diffusion model during validation, Mean Absolute Error (MAE) for the regressor model during validation and Validity and Uniqueness when sampling molecules for testing. Lower values of NLL and MAE indicate better performance, while higher values of Validity and Uniqueness are desirable. Validity is measured using RDKit sanitization and Uniqueness is assessed over 1,000 generated molecules. Furthermore, to better compare the performance between CL (\(M_{\text{central}}\)) and FL (\(M_{\text{federated}}\)), we calculate the absolute percent difference as follows:
\begin{equation}
P = \left( \frac{\left| M_{\text{central}} - M_{\text{federated}} \right|}{\frac{M_{\text{central}} + M_{\text{federated}}}{2}} \right) \times 100\%
\end{equation}

The results of our experiments are summarized in Table \ref{tab1}. The centralized learning model achieves an NLL of 68.45, an MAE of 0.6932, a Validity of 0.9600, and a Uniqueness of 0.9958. In comparison, the federated learning model achieves an NLL of 70.58, an MAE of 0.7026, a Validity of 0.9560, and a Uniqueness of 0.9989. The percent differences between the two methods are 3.11\% for NLL, 1.36\% for MAE, 0.42\% for Validity, and 0.31\% for Uniqueness. These results indicate that the performance of the FL model is highly comparable to that of the CL model, suggesting that a model can be trained across different data sites with minimal loss of performance. 

\begin{figure}[t]
\includegraphics[width=\textwidth]{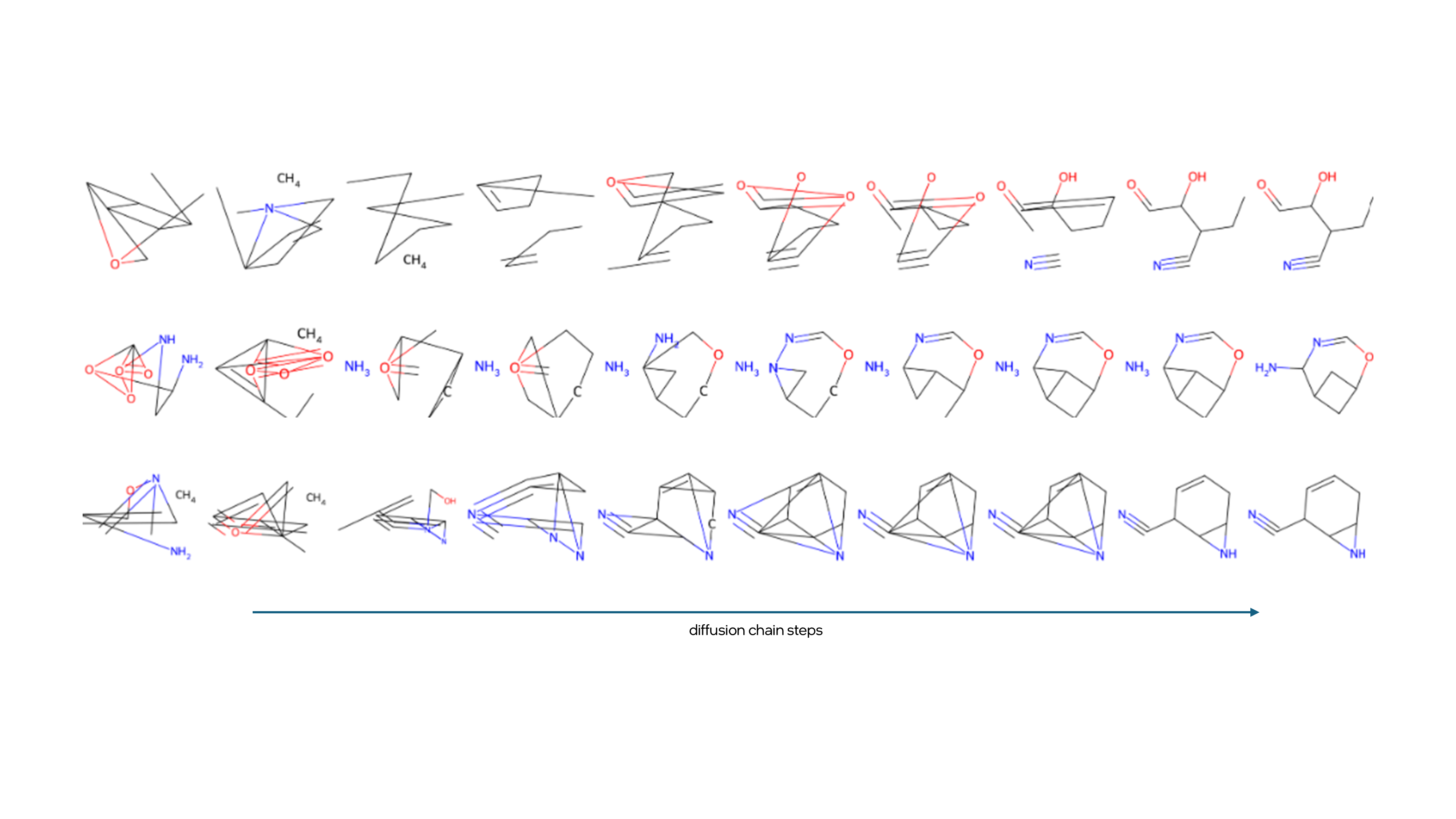}
\caption{\small Example of three molecules generated by the federated model. The diffusion chains illustrate generation from a noisy graph to a plausible molecule} \label{fig-molecules}
\end{figure}

To further illustrate the comparability of the two learning paradigms, we present training and validation curves for both the regressor model and the denoising model in Figures \ref{fig:training-curves} and \ref{fig:validation-curves}, respectively. These curves show that the federated model's performance is nearly identical to that of the centralized model during training, with both models converging to similar levels of accuracy and loss. Additionally, we provide a graph showing the progress of Uniqueness and Validity over training rounds for the full guidance model, which combines the denoising model with the regressor model in Figure \ref{fig:sampling-metrics}. These graphs demonstrate that the federated learning model maintains high levels of Uniqueness and Validity throughout the training process, closely mirroring the performance of the centralized learning model. Figure \ref{fig-molecules} illustrates three unique molecules generated from the federated model and the diffusion chain steps to achieve the final, valid molecule, demonstrating the feasibility of using a federated model for molecule generation.

\section{Conclusion \& Future Directions}

In this work, we presented a federated discrete denoising diffusion model molecular generation by integrating the DiGress model \cite{vignac2023digress} into the OpenFL framework \cite{Foley2022-vl}. Our experimental results demonstrate that federated learning can achieve performance comparable to centralized learning in the task of AI-driven molecular generation. Empirical results for NLL, MAE, Validity, and Uniqueness show that federated learning can match the performance of centralized learning while offering significant privacy and security benefits. These findings emphasize the potential of federated learning to facilitate collaborative research and development in the pharmaceutical industry, enabling the generation of novel drug candidates while preserving data privacy and allowing individual companies and organizations to maintain their intellectual property.

Our work highlights the utility of federated learning in the drug design process. It provides a viable alternative to centralized learning that leverages decentralized data sources without compromising performance. This approach paves the way for more secure and collaborative AI-driven drug discovery efforts, ultimately accelerating the development of new therapeutics. 

While our study demonstrates significant promise, there are exciting opportunities for further enhancement. For instance, adopting a scaffold split instead of a randomized data split could mitigate the risk of data leakage in the training set, ensuring even greater data integrity. Additionally, while the QM9 dataset's smaller molecules have provided valuable insights, it is also easier to achieve high levels of uniqueness. Therefore, testing our approach on larger and more complex molecular datasets such as GuacaMol \cite{Brown2019-ou} would help validate its robustness and generalizability. Furthermore, as the scale and complexity of data continue to grow, leveraging more sophisticated federated aggregation algorithms, which OpenFL supports, may become necessary. Future research directions include integrating additional advanced privacy-preserving techniques and federated aggregation algorithms, expanding the number of collaborators, and additional studies on larger molecular datasets, such as GuacaMol, with more robust data splitting techniques. Addressing challenges such as heterogeneous data distributions and fostering collaborations across the pharmaceutical industry will be crucial for the successful deployment of federated learning in real-world scenarios. By advancing this technology, we can unlock new opportunities for innovation and collaboration toward discovering state-of-the-art therapeutics.

\bibliographystyle{splncs04} 
\bibliography{main} 
\end{document}